\documentclass[letterpaper]{article}
\usepackage{proceed2e}
\usepackage[margin=1in]{geometry}

\usepackage{times}

\usepackage[pdftex]{graphicx} %
\usepackage{amsmath, amssymb,amsthm}
\usepackage{color}

\newtheorem{theorem}{Theorem}

\usepackage{dblfloatfix}
\usepackage[linesnumbered,ruled]{algorithm2e}
\usepackage[round]{natbib}   %
\usepackage{subcaption}
\usepackage{natbib}
\setcitestyle{authoryear,open={[},close={]}}

\title{Parallelizable sparse inverse formulation Gaussian processes (SpInGP)}

\author{ {\bf Alexander Grigorievskiy\thanks{\texttt{alexander.grigorevskiy@aalto.fi} \newline Alexander thanks \textbf{EIT Digital} for funding his research visit to Sheffile University}} \\
Aalto University, Finland \\
\And
{\bf Neil Lawrence} \\
University of Sheffield \\ Amazon Research, Cambridge, UK\\
\And
{\bf Simo S\"{a}rkk\"{a}}   \\
Aalto University, Finland\\
}

\graphicspath{ {../Images/} }
\begin{document}

\maketitle

\begin{abstract}
We propose a parallelizable sparse inverse formulation Gaussian process (SpInGP) for temporal models. It uses a sparse precision GP formulation and sparse matrix routines to speed up the computations. Due to the state-space formulation used in the algorithm, the time complexity of the basic SpInGP is linear, and because all the computations are parallelizable, the parallel form of the algorithm is sublinear in the number of data points. We provide example algorithms to implement the sparse matrix routines and experimentally test the method using both simulated and real data.
\end{abstract}

\section{Introduction}
	
Gaussian processes (GPs) are prominent modelling techniques in machine learning~\citep{rasmussen_2005}, robotics, signal processing, and many others fields. The advantages of Gaussian Process models are: analytic tractability in the basic case and probabilistic formulation which offers handling of uncertainties.  The computational complexity of standard GP regression is $O(N^3)$ where $N$ is the number of data points. It makes the inference impractical for sufficiently large datasets. For example, modelling of a dataset with 10,000--100,000 points on a modern laptop becomes slow, especially if we want to estimate hyperparameters~\citep{rasmussen_2005}. However, the computational complexity of a standard GP does not depend on the dimensionality of the input variables. 

Several methods have been developed to reduce the computational complexity of GP regression. These include inducing point sparse methods~\citep{quinonero2005}, which assume the existence of a smaller number $m$ of inducing inputs and inference is done using these inputs instead of the original inputs, the overall complexity of this approximation is $O(m^2N)$ -- independently of the dimensionality of inputs. Although the scaling is also linear in $N$, these methods are approximations with no way to quantify their imprecision. %

In this paper, we consider temporal Gaussian processes, that is, processes which depend only on time. Hence, the input space is 1-dimensional (1-D). Temporal GPs are frequently used e. g. for trajectory estimation problems in robotics~\citep{anderson15} and signal processing. For such 1-D input GPs there exist another method to reduce the computational complexity~\citep{hartikainen_2010}. This method allows to convert temporal GP inference to the linear state-space (SS) model and to perform model inferences by using Kalman filters and Raugh--Tung--Striebel (RTS) smoothers \citep{sarkka_book}. This method reduces the computational complexity of temporal GP to $O(N)$ where $N$ is the number of time points.

Even though the method described in the previous paragraph scales as $O(N)$ we are interested in speeding up the algorithm even more. The computational complexity cannot, in general, be lower than $O(N)$ because the inference algorithm has to read every data point at least once. However, with finite $N$, by using parallelization we can indeed obtain sublinear computational complexity in ``weak sense'': provided that the original algorithm is $O(N)$, and it is parallelizable, the solution can be obtained with time complexity which is strictly less than linear. Although the state-space formulation together with Kalman filters and RTS smoothers provide the means to obtain linear time complexity, the Kalman filter and RTS smoother are inherently \emph{sequential} algorithms, which makes them difficult to parallelize. 

In this paper, we show how the state-space formulation can be used to construct linear-time inference algorithms which use the sparseness property of precision matrices of Markovian processes. The advantage of these algorithms is that they are parallelizable and hence allow for sublinear time complexity (in the above weak sense). Similar ideas have also been earlier proposed in robotics literature \citep{anderson15}. We discuss practical algorithms for implementing the required sparse matrix operations and test the proposed method using both simulated and real data.

\section{Sparse Precision Formulation of Gaussian Process Regression}

\subsection{State-Space Formulation and Sparseness of Precision matrix}

It has been shown before~\citep{hartikainen_2010}, that a wide class of temporal Gaussian processes can be represented in a state-space form:
\begin{equation}\label{eq:ss1}
\begin{aligned}
& \qquad \mathbf{z}_n = A_{n-1} \mathbf{z}_{n-1} + \mathbf{q}_n \quad \text{(dynamic model)}, \\
& \qquad y_n = H \mathbf{z_n} + \epsilon_n \quad \text{(measurement model)}
\end{aligned},
\end{equation}
where noise terms are $\mathbf{q}_n \sim \mathcal{N}(0, Q_n)$, $\epsilon_n \sim \mathcal{N}(0,\sigma_n^2)$, and the initial state is $\mathbf{z}_0 \sim \mathcal{N}(0,P_0)$.

It is assumed that $y_n$ (scalar) are the observed values of the random process. The noise terms $\mathbf{q}_n$ and $\epsilon_n$ are Gaussian. The exact form of matrices $A_n, Q_n, H $ and $P_0$ and the dimensionality of the state vector -- $b$ depend on the covariance matrix of the corresponding GP. Some covariance functions have an exact representation in a state-space form such as the Mat{\'e}rn family~\citep{hartikainen_2010}, while some others have only approximative representations, such as the exponentiated quadratic (EQ)~\citep{hartikainen_2010} (it is also called radial basis function or RBF kernel). It is important to emphasize that these approximations can be made arbitrarily small in a controllable way because of a uniform convergence of approximate covariance function to the true covariance. This is in contrast with e.g. sparse GPs where there is no control of the quality of approximation.

Usually, the state-space form~\eqref{eq:ss1} is obtained first by deriving a stochastic differential equation (SDE) which corresponds to GP regression with the appropriate covariance function, and then by discretizing this SDE. The transition matrices $A_n$ and $Q_n$ actually depend on the time intervals between the consecutive measurements so we can write $A_n = A[t_{n+1} - t_{n}] = A[\Delta t_n]$ and $Q_n = Q[\Delta t_n]$. Matrix $H$ typically is fixed and does not depend on $\Delta t_n$. 

For instance, the state-space model for the Mat{\'e}rn$(\nu = \frac{3}{2})$ covariance function $k_{\nu=3/2}(t) = (1 + \frac{\sqrt{3} \Delta t}{l}) \exp(-\frac{\sqrt{3} \Delta t}{l}) $ is:

\begin{equation}\label{eq:ss_matern}
\begin{aligned}
& A_{n-1} = \mathtt{expm} \left( \begin{bmatrix} 0 & 1 \\ -\phi^2 & -2 \phi \end{bmatrix} \Delta t_n \right), \\
& P_0 =  \begin{bmatrix} 1 & 0 \\ 0 & \frac{3}{l^2} \end{bmatrix}, \qquad H= \begin{bmatrix} 1 & 0 \end{bmatrix} \\
& Q_n = P_0 - A_{n-1} P_0 A_{n-1}^T \\
\end{aligned},
\end{equation}

where $\mathtt{expm}$ denotes matrix exponent and $\phi = \frac{\sqrt{3} }{l}$.

We do not consider here how exactly the state-space form is obtained, and interested reader is guided to the references \citep{hartikainen_2010} and later works on this topics. However, regardless of the method we use, we have the following group property for the transition matrix:
\begin{equation}
A[\Delta t_n] A[\Delta t_k] = A[\Delta t_n + \Delta t_k].
\end{equation}
Having a representation~\eqref{eq:ss1} we can consider the vector of $\bar{\mathbf{z}}=[\mathbf{z}_0^{\top}, \mathbf{z}_1^{\top}, \cdots \mathbf{z}_N^{\top}]^{\top}$ which consist of individual vectors $\mathbf{z}_i$ stacked vertically. The distribution of this vector is Gaussian because the whole model~\eqref{eq:ss1} is Gaussian. It has been shown in~\citep{grigorievskiy_2016} and in different notation in~\citep{anderson15} that the covariance of $\bar{\mathbf{z}}$ equals:

\begin{equation}\label{eq:k1}
\mathcal{K}(\mathbf{t}, \mathbf{t}) = \mathbf{A} \mathbf{Q} \mathbf{A}^{\top}.
\end{equation}
In this expression the matrix $\mathbf{A}$ is equal to:
\begin{equation}
\resizebox{0.75\hsize}{!}{$
\mathbf{A} = \begin{bmatrix}
I & 0 & 0 & \cdots & 0\\
A[\Delta t_{1}] & I & 0 &  & 0\\
A[\Delta t_{1} + \Delta t_{2}] & A[\Delta t_{2}] & I & \cdots & 0\\
\vdots & & \vdots & \ddots & \vdots\\
A[ \sum\limits_{i=1}^{N} \Delta t_{i} ] & A[ \sum\limits_{i=2}^{N} \Delta t_{i} ] & A[ \sum\limits_{i=3}^{N} \Delta t_{i} ] & \cdots & I\\
\end{bmatrix}.$}
\end{equation}
The matrix $\mathbf{Q}$ is
\begin{equation}\label{eq:q_matrix}
\resizebox{0.50\hsize}{!}{$
\mathbf{Q} = \begin{bmatrix} 
P_0 & 0 & 0 & \cdots & 0\\
0  & Q_1 & 0 &  & 0\\
0 & 0 & Q_2 & \cdots & 0 \\
\vdots & &  \vdots & \ddots & \vdots \\
0 & 0  & 0 & \cdots & Q_N
\end{bmatrix}.$}
\end{equation}

There exist the following theorem (e.g. \citep{anderson15}) which shows the sparseness of the inverse of the covariance matrix: 

\begin{equation}\label{eq:hh1}
\mathcal{K}^{-1} = \mathbf{A}^{-T} \mathbf{Q}^{-1} \mathbf{A}^{-1}.
\end{equation}.

\begin{theorem}\label{trm:1}
The inverse of the kernel matrix $\mathcal{K}$ from Eq.~\eqref{eq:k1} is a block-tridiagonal (BTD) and therefore is a sparse matrix.
\end{theorem}

The above result can be obtained by noting that the $ \mathbf{Q}^{-1}$ is block diagonal (denote the block size as $b$) and that

\begin{equation}
\resizebox{0.67\hsize}{!}{$
\mathbf{A}^{-1} = \begin{bmatrix}
I & 0 & 0 & \cdots & 0\\
-A[\Delta t_{1}] & I & 0 &  & 0\\
0 & -A[\Delta t_{2}] & I & \cdots & 0\\
\vdots & & \vdots & \ddots & \vdots\\
 0 & 0 & 0 & -A[\Delta t_{N}] & I\\
\end{bmatrix}.$}
\end{equation}

It is easy to find the covariance of observation vector $\mathbf{y} = [y_1, y_2, \cdots y_N]^{\top}$. In the state-space model~\eqref{eq:ss1} observations $y_i$ are just linear transformations of state vectors $\mathbf{z}_i$, hence according to the properties of linear transformation of Gaussian variables, the covariance matrix in question is:

\begin{equation}\label{eq:k2}
\begin{split}
G\,\mathcal{K}\, G^{\top} + I_N \sigma_n^2,
\end{split}
\end{equation}

where $G = (I_N \otimes H)$.

The symbol $\otimes$ denotes the Kronecker product. Note, that in state-space model Eq.~\eqref{eq:ss1}, $H$ is a row vector because $y_i$ are 1-dimensional. The summand $I_N \sigma_n^2$ correspond to the observational noise can be ignored when we refer to the covariance matrix of the model. Therefore, we see that the covariance matrix of $\mathbf{y}$ has the inner part $\mathcal{K}$ which inverse is sparse (block-tridiagonal). This property can be used for computational convenience via \textit{matrix inversion lemma}.

\subsection{Sparse Precision Gaussian Process Regression}\label{sec:mean_and_var}

In this section, we consider sparse inverse (SpIn) (which we also call sparse precision) Gaussian process regression and computational subproblems related to it.
Let's look at temporal GP with redefined covariance matrix in Eq.~\eqref{eq:k2}. The only difference to the standard GP is the covariance matrix. Since the state-space model might be an exact or approximate expression of GPR we can write:
\begin{equation}
K(\mathbf{t},\mathbf{t}) \approxeq G\,\mathcal{K}(\mathbf{t},\mathbf{t})\, G^{\top}.
\end{equation}.

The predicted mean value of a GP at $M$ new time points $\mathbf{t}^{\star} = [t_1^{\star}, t_2^{\star}, \cdots, t_M^{\star}]$ is:
\begin{equation}\label{eq:mean2}
m(\mathbf{t}^{\star}) = G_M \mathcal{K}(\mathbf{t}^{\star},\mathbf{t}) G^{\top} [G \mathcal{K}(\mathbf{t},\mathbf{t}) G^{\top} + \Sigma ]^{-1} \mathbf{y},
\end{equation}

where we have denoted $\Sigma = \sigma_n^2 I$. One way to express the mean through the inverse of the covariance $\mathcal{K}^{-1}(\mathbf{t},\mathbf{t})$ is to apply the matrix inversion lemma:
\begin{equation}
\begin{split}
&[G \mathcal{K}(\mathbf{t},\mathbf{t}) G^{\top} + \Sigma]^{-1} \\
&= \Sigma^{-1}
- \Sigma^{-1} G [\mathcal{K}(\mathbf{t},\mathbf{t})^{-1} + G^{\top} \Sigma^{-1} G]^{-1} G^{\top} \Sigma^{-1}
\end{split}.
\nonumber
\end{equation}

After substituting this expression into Eq.~\eqref{eq:mean2} we get:

\begin{equation}\label{eq:mean3}
\begin{split}
  m(\mathbf{t}^{\star}) 
  &= G_M \mathcal{K}(\mathbf{t}^{\star},\mathbf{t}) G^{\top} [\Sigma^{-1} \mathbf{y}
- \Sigma^{-1} G \times \\
&\underbrace{[\mathcal{K}(\mathbf{t},\mathbf{t})^{-1}+G^{\top} \Sigma^{-1} G]^{-1} (G^{\top} \Sigma^{-1} \, \mathbf{y})]}_{\text{Computational subproblem 1}} \\
\end{split},
\end{equation}

where $G_M = (I_M \otimes H)$.  The inverse of the matrix $\mathcal{K}(\mathbf{t},\mathbf{t})$ is available analytically from Theorem~\ref{trm:1}. It is a block-tridiagonal (BTD) matrix. The matrix $G^{\top} \Sigma^{-1} G$ is a block-diagonal matrix which follows from:

\begin{equation}
G^{\top} \Sigma^{-1} G
= \sigma_n (I_N \otimes H^{\top}) (I_N \otimes H) = \sigma_n^2 (I_N \otimes H^{\top} H).
\nonumber
\end{equation}

Thus, the main computational task in finding the mean value for new time points is solving block-tridiagonal system with matrix $[\mathcal{K}(\mathbf{t},\mathbf{t})^{-1} + G^{\top} \Sigma^{-1} G]^{-1}$ and right-hand side $(G^{\top} \Sigma^{-1} \, \mathbf{y})$.
However, note that even though the matrix $\mathcal{K}^{-1}(\mathbf{t},\mathbf{t})$ is sparse (block-tridiagonal), the inverse matrix is dense. So, the matrix $\mathcal{K}(\mathbf{t}^{\star},\mathbf{t})$ is dense as well.

The computational complexity of the above formula is $O(MN)$, however if $M$ is large then the matrix $\mathcal{K}(\mathbf{t}^{\star},\mathbf{t})$ may not fit into computer memory because it is dense. Of course, it is possible to apply~\eqref{eq:mean3} in batches taking each time a small number of new time points, but this is cumbersome in implementation. 

There is another formulation of computing the GP mean with the same computational complexity. We combine the training and test points in one vector $T = [\mathbf{t}^{\star}; \mathbf{t}]$ of the size $L=M+N$ and consider the GP covariance formula for the full covariance $K(T,T)$. Note, that according to the previous discussion $K(T,T) = G \mathcal{K}(T,T) G^{\top}$ and we try to express the covariance $\mathcal{K}(T,T)$ through the inverse $\mathcal{K}^{-1}(T,T)$ which is sparse. Then the mean formula for $T$ is:

\begin{equation}\label{eq:mean4}
m(T) = G_L \; \mathcal{K}(T,T) \; G_L^T\left[ G_L \; \mathcal{K}(T,T) \; G_L^T+ \Sigma^{\prime} \right]^{-1} \mathbf{y}^{\prime}.
\end{equation}

Here the $G_L$ by analogy equals $(I_L \otimes H)$, and $\Sigma^{\prime}$ contains $\sigma_n^2$ on those
diagonal positions which correspond to training points and $\infty$ on those which correspond to new points (infinity must be understood in the limit sense). The vector $\mathbf{y}^{\prime}$ is similarly augmented with zeros. 

Then by applying the following matrix identity:
\begin{equation}\label{eq:mean5}\nonumber
A^{-1}B [D - C A^{-1} B]^{-1} = [A - BD^{-1}C]^{-1} B D^{-1},
\end{equation}

we arrive to:
\begin{equation}\label{eq:mean6}
m(T) =  G_L \underbrace{ [ \mathcal{K}^{-1} + G^T (\Sigma^{\prime})^{-1} G ]^{-1} G_L^{\top} (\Sigma^{\prime})^{-1} \mathbf{y}^{\prime} }_{\text{Computational subproblem 1}}.
\end{equation}

Let's consider the variance computation. If we apply straightforwardly the matrix inversion lemma to the GP variance formula,  the result is:

\begin{equation}
\nonumber
\begin{aligned}
&S(\mathbf{t}^{\star},\mathbf{t}^{\star})
= G_M \mathcal{K}(\mathbf{t}^{\star},\mathbf{t}^{\star}) G_M^{\top} - G_M \mathcal{K}(\mathbf{t}^{\star},\mathbf{t}) G^{\top} \\
&\qquad \times [G \mathcal{K}(\mathbf{t},\mathbf{t}) G^{\top} + \Sigma^{\prime}]^{-1} G \mathcal{K}(\mathbf{t},\mathbf{t}^{\star}) G_M^{\top}.
\end{aligned}
\end{equation}

In this expression the computational complexity is not less than $O(N^3)$ and the matrix $\mathcal{K}(\mathbf{t}^{\star},\mathbf{t}^{\star})$ is dense. This is computationally prohibitive for the large datasets. By using the similar procedure as for the mean computations we derive:

\begin{equation}\label{eq:k3}
S(T,T)=G_L \underbrace{[\mathcal{K}^{-1} + G_L^{\top} (\Sigma^{\prime})^{-1} G_L]^{-1} G_L^{\top}}_{\text{Computational subproblem 2}}.
\end{equation}

Typically we are interested only in the diagonal of the covariance matrix~\eqref{eq:k3}, therefore not all the elements need to be computed. Section~\ref{sec:comp_subprob} describes in more details this and other computational subproblems.

\subsection{Marginal Likelihood}

The formula for the marginal likelihood of GP is:
\begin{equation}\label{eq:ml2}
\begin{split}
\log p(\mathbf{y}| \mathbf{t}) = \underbrace{ -\frac{1}{2} \mathbf{y}^{\top}[K + \Sigma]^{-1} \mathbf{y}}_{\text{data fit term: ML\_1}} &\underbrace{-\frac{1}{2} \log \det[K + \Sigma]}_{\text{determinant term: ML\_2}}\\ &
- \frac{n}{2} \log 2\pi
\end{split}
\end{equation}
Computation of the \textit{data fit term} is very similar to the mean computation in Section~\ref{sec:mean_and_var}. For computing the \textit{determinant term} the \textit{determinant Inverse Lemma} must be applied. It allows to express the determinant computation as:

\begin{equation}\label{eq:det_1}
\begin{split}
&\det[K + \Sigma] = \det[G \mathcal{K} G^{\top} + \Sigma] = \\
&= \underbrace{\det[ \mathcal{K}^{-1} + G^{\top} \Sigma^{-1} G]}_{\text{Computational subproblem 3}} \det[\mathcal{K}] \det[\Sigma]
\end{split}
\end{equation}

We assume that $\Sigma$ is diagonal, then in the formula above $\det[\Sigma]$ is easy to compute: $\det[\Sigma]=(\sigma_{n}^{2})^{N}$ ($N$ - number of data points). $\det[\mathcal{K}] = 1 / \det[\mathcal{K}^{-1}]$, so we need to know how to compute the determinants of BTD matrices. This constitutes the third computational problem to be addressed.

\subsection{Marginal Likelihood Derivatives Calculation}

Taking into account GP mean calculation and the marginal likelihood form in Eq~\eqref{eq:ml2} we can write:

\begin{equation}
\begin{aligned}
&\log p(\mathbf{y}| \mathbf{t})= \\
&=-\frac{1}{2} \mathbf{y}^{\top} \left( \Sigma^{-1} - \Sigma^{-1} G [\mathcal{K}^{-1} + G^{\top}\Sigma^{-1}G]^{-1} \right) \mathbf{y} \\
&-\frac{1}{2} \Bigg( \log \det[ \mathcal{K}^{-1} + G^{\top} \Sigma^{-1} G ] - \log\det[\mathcal{K}^{-1}] \\
&+ \log\det[\Sigma^{-1}] \Bigg) - \frac{N}{2} \log 2\pi.
\end{aligned}
\end{equation}
Consider first the derivatives of the data fit term. Assume that $\theta \neq \sigma_n$ (variance of noise). So, $\theta$ is a parameter of a covariance matrix, then:
\begin{equation}
\begin{split}
\frac{\partial \text{ML\_1}}{ \partial \theta} &= \left\{ \text{using:} \;\; \frac{\partial M^{-1}}{ \partial \theta}  = -M^{-1} \frac{\partial M}{ \partial \theta} M^{-1} \right\}= \\
&-\frac{1}{2} \mathbf{y}^{\top} \Big( \Sigma^{-1} G [ \mathcal{K}^{-1} + G^{\top} \Sigma^{-1} G]^{-1} \frac{\partial \mathcal{K}^{-1} }{ \partial \theta} \times \\
& \qquad \quad \times [ \mathcal{K}^{-1} + G^{\top} \Sigma^{-1} G]^{-1} G^{\top} \Sigma^{-1} \Big) \mathbf{y}.
\end{split}
\end{equation}

This is quite straightforward (analogously to mean computation) to compute if we know $\frac{\partial \mathcal{K}^{-1} }{ \partial \theta}$. This is computable using Theorem~\ref{trm:1} and expression for the derivative of the inverse. If $\theta = \sigma_n$ the derivative is computed similarly.

Assuming again that the $\theta$ is a parameter of the covariance matrix, the derivative of the determinant is:
\begin{equation}
\frac{\partial }{ \partial \theta} \left\{ -\frac{1}{2} \left( \log\det[ \mathcal{K}^{-1} + G^{\top} \Sigma^{-1} G ] - \log\det[\mathcal{K}^{-1}] \right) \right\}.
\end{equation}

Let us consider how to compute the first $\mathit{\log\det}$ since computing the second one is analogous.

\begin{equation}\label{eq:dd1}
\begin{split}
\frac{\partial }{ \partial \theta} &\left\{ \log\det[\mathcal{K}^{-1}+ G^{\top} \Sigma^{-1} G] \right\} =\\
&= \underbrace{Tr\left[ [\mathcal{K}^{-1}+ G^{\top} \Sigma^{-1} G]^{-1} \frac{\partial [\mathcal{K}^{-1}+ G^{\top} \Sigma^{-1} G]}{ \partial \theta} \right]}_{\text{Computational subproblem 4}}
\end{split}.
\end{equation}

The question is how to efficiently compute this trace? Briefly denote $\mathbf{K} = \mathcal{K}^{-1}+ G^{\top} \Sigma^{-1} G$. Although the matrix $\mathbf{K}$ is sparse, the matrix $\mathbf{K}^{-1}$ is dense and can't be obtained explicitly. One can think of computing sparse Cholesky decomposition of $\mathbf{K}^{-1}$ then solving linear system with right-hand side (rhs) $\frac{\partial \mathbf{K}^{-1}}{ \partial \theta}$ (rhs is a matrix). After solving the system it is trivial to compute the trace. However, this approach faces the problem that the solution of the linear system is dense and hence can't be stored.

To avoid this problem we can think of performing sparse Cholesky decomposition of both $\mathbf{K}^{-1}$ and $\frac{\partial \mathbf{K}^{-1}}{ \partial \theta}$ in Eq.~\eqref{eq:dd1} in order to deal with triangular matrices with the intention to compute the trace without dealing with large matrices. However, after careful investigation, we conclude that this approach is infeasible. We consider the efficient solution to this problem in the next section.

\section{Computational Subproblems and their Solutions}\label{sec:comp_subprob}

\subsection{Overview of Computational Subproblems}\label{sec:comp_probl}

Let's consider one by one computational subproblems defined earlier. They all involve solving numerical problems with \textbf{symmetric} block-tridiagonal (BTD) matrix
e.g. in Eq.~\eqref{eq:mean3}.

The mean computation in SpInGP formulation requires solving \textbf{computational subproblem 1} which is emphasized in Eq.~\eqref{eq:mean3}. It is a block-tridiagonal linear system of equations:
\begin{equation}\label{eq:comp_1}
\resizebox{0.65\hsize}{!}{$
\left[ \begin{matrix} \boxed{?} \\
\boxed{?} \\ \vdots \\ \boxed{?} \end{matrix} \right]
= \begin{bmatrix} 
A_1 & B_1 & \cdots & 0\\
C_1  & A_2 &  & 0\\
 \vdots & \vdots & \ddots & \vdots \\
 0 & 0 & \cdots & A_n \\
\end{bmatrix}^{-1}  
\left[ \begin{matrix} x \\
x \\ \vdots \\ x \end{matrix} \right]$}
\end{equation}

This is a standard subproblem which can be solved by classical algorithms. There is a Thomas algorithm which is sequential. It performs block LU factorization of the given matrix. Parallel version has been developed as well. Unfortunately, block matrix algorithms are rarely found in numerical libraries.

The \textbf{computational subproblem 3} in Eq.~\eqref{eq:det_1} involves computing the determinant of the same symmetric block-tridiagonal matrix. In general, any direct (not iterative) solver performs some version of LU (Cholesky in the symmetric case) decomposition, therefore usually subproblem 3 is solved simultaneously with subproblem 1.

\begin{equation}\label{eq:comp_2}
\resizebox{0.85\hsize}{!}{$
\begin{bmatrix} 
\boxed{?} & x & \cdots & x \\
x  & \boxed{?} &  & x \\
 \vdots & \vdots & \ddots & \vdots \\
 x & x & \cdots & \boxed{?} \\
\end{bmatrix}
= \begin{bmatrix} 
A_1 & B_1 & \cdots & 0\\
C_1  & A_2 &  & 0\\
 \vdots & \vdots & \ddots & \vdots \\
 0 & 0 & \cdots & A_n \\
\end{bmatrix}^{-1}  \times
\begin{bmatrix} 
X & X & \cdots & 0\\
X  & X &  & 0\\
 \vdots & \vdots & \ddots & \vdots \\
 0 & 0 & \cdots & X \\
\end{bmatrix}$}
\end{equation}
Alternatively, we can tackle subproblems 1 and 3 by using general band matrix solvers or sparse solvers. There are several general purpose \textbf{direct} sparse solvers available: \textit{Cholmod}, MUMPS, PARDISO etc. The restriction that the solver must be direct follows from the need to compute the determinant in subproblem 3. 

The \textbf{computational subproblem 2} in Eq.~\eqref{eq:k3} is a less general form of \textbf{computational problem 4}. The scheme of the subproblem 4 is demonstrated in Eq.~\eqref{eq:comp_2}.
On this scheme small $x$ means any single element and $X$ any block. In short, BTD matrix is inverted and the right-hand side (rhs) is also a BTD matrix. Right-hand side blocks do not have to be square, the only requirement is that the dimensions match. Since the inversion of the block-tridiagonal matrix is a dense matrix the solution of this problem is also a dense matrix. However, we are interested not in the whole solution but only in the diagonal of it.

It can be noted, that subproblem 4 can be solved by computing the BTD part of the inverse of the matrix and then multiplied by the right-hand side. This is true since the right-hand side is a BTD matrix. Hence, only the BTD part of the inverse is needed to compute the required diagonal. Computing only some elements in an inverse matrix is called \textit{selective inversion}. Some direct sparse solvers implement selective inversion parallel algorithm. Hence, subproblem 4 can also be solved by a general sparse solvers. However, developing the specialized numerical algorithms for BTD matrices may bring better performance~\citep{petersen2009}.

\section{Experiments}

\begin{figure*}[h]
    \centering
    \begin{subfigure}[h]{0.33\textwidth}
        \centering
        \includegraphics[width=0.8\columnwidth]{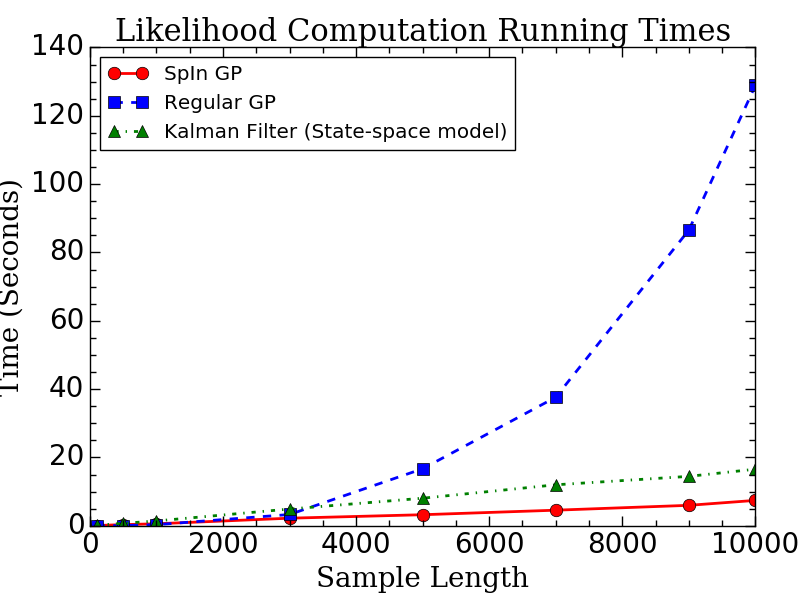}
        \caption{Scaling of SpInGP w.r.t. number of data points. Marginal likelihood and its derivatives are being computed.}\label{fig:1}
    \end{subfigure}%
    ~ 
    \begin{subfigure}[h]{0.33\textwidth}
        \centering
        \includegraphics[width=0.8\columnwidth]{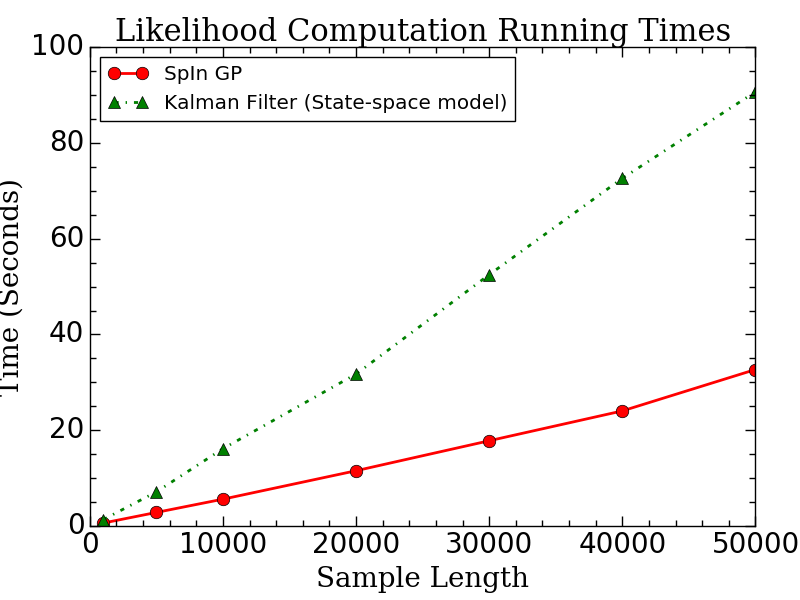}
        \caption{Scaling of SpInGP and KF w.r.t. number of data points. Marginal likelihood and its derivatives are being computed.}\label{fig:2}
    \end{subfigure}%
    ~ 
    \begin{subfigure}[h]{0.33\textwidth}
        \centering
        \includegraphics[width=0.8\columnwidth]{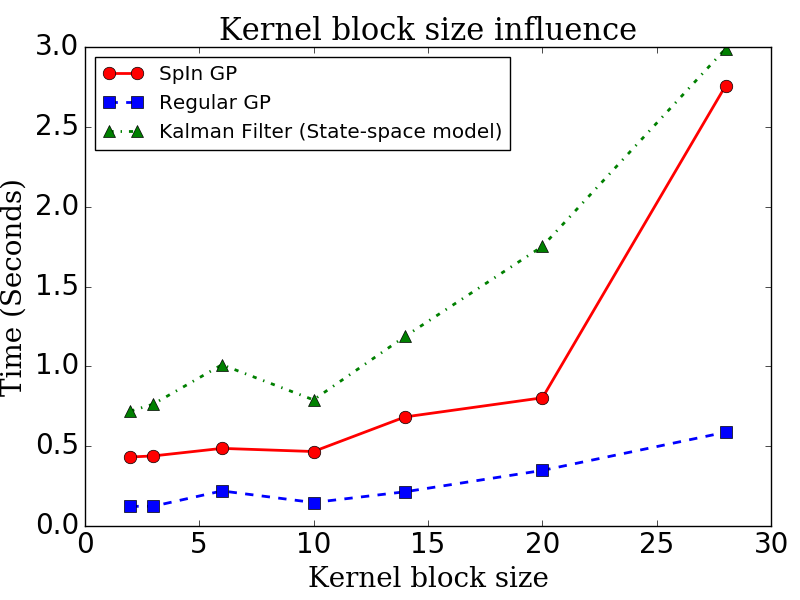}
        \caption{Scaling of SpInGP w.r.t block size of precision matrix. 1000 data points. Marginal likelihood and its derivatives are being computed.}\label{fig:3}
    \end{subfigure}
    \caption{Scaling of SpIn GP}\label{fig:6}
\end{figure*}

\subsection{Implementation}

The sequential algorithms for SpInGP based on Thomas algorithm have been implemented\footnote{Source code available at: (added in the final version)}. The implementation of parallel versions has been postponed for the future because it requires fine-tuning (in the case of using available libraries) or substantial efforts if implemented from scratch.

The implementations are done in Python environment using the \textit{Numpy} and \textit{Scipy} numerical libraries. The code is also integrated with the \textit{GPy}~\citep{gpy2014} library where many state-space kernels and a rich set of GP models are implemented. Results are obtained on a regular laptop computer with Intel Core i7 CPU @ 2.00GHz × 8 and with 8 Gb of RAM.

\subsection{Simulated Data Experiments}

We have generated an artificial data which consist of two sinusoids immersed into Gaussian noise. The speed of computing marginal log likelihood (MLL) along with its derivatives are presented in the Fig.~\ref{fig:1}. SpInGP is compared with state-space form of temporal GP regression (also implemented in GPy). The inference in state-space model is done by Kalman filtering, so we call it also KF model. The block size is $b=12$ which correspond to Mat{\'e}rn$(\nu=\frac{3}{2})$ plus EQ (RBF) with 10-th order approximation. The same test but for a larger number of data points is done~(Fig.~\ref{fig:2}) to verify linear memory consumption.

As expected the SpInGP and KF solution scales linearly with increasing number of data points while standard GP scales cubically. The SpInGP is faster than KF here, but it is only because of implementation details. 

\begin{figure*}[h]
    \centering
    \begin{subfigure}[h]{0.33\textwidth}
        \centering
        \includegraphics[width=0.8\columnwidth]{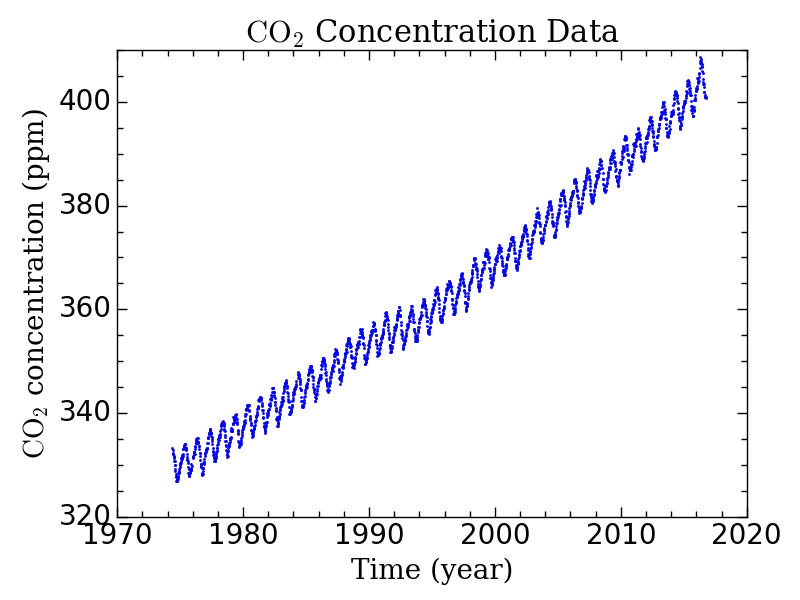}
        \caption{$CO_2$-concentration dataset. 2195 data points}\label{fig:5a}
    \end{subfigure}%
    ~ 
    \begin{subfigure}[h]{0.33\textwidth}
        \centering
        \includegraphics[width=0.8\columnwidth]{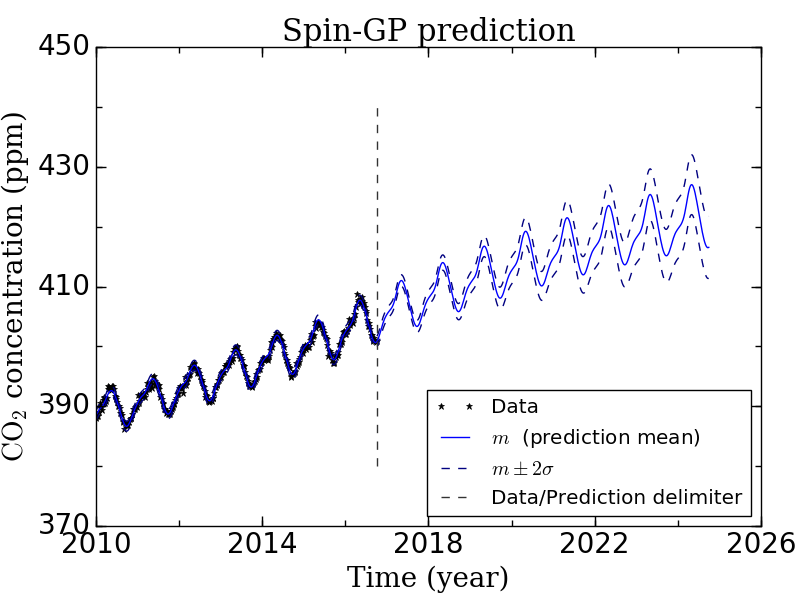}
        \caption{SpInGP 8 years forecast}\label{fig:5b}
    \end{subfigure}%
    ~ 
    \begin{subfigure}[h]{0.33\textwidth}
        \centering
        \includegraphics[width=0.8\columnwidth]{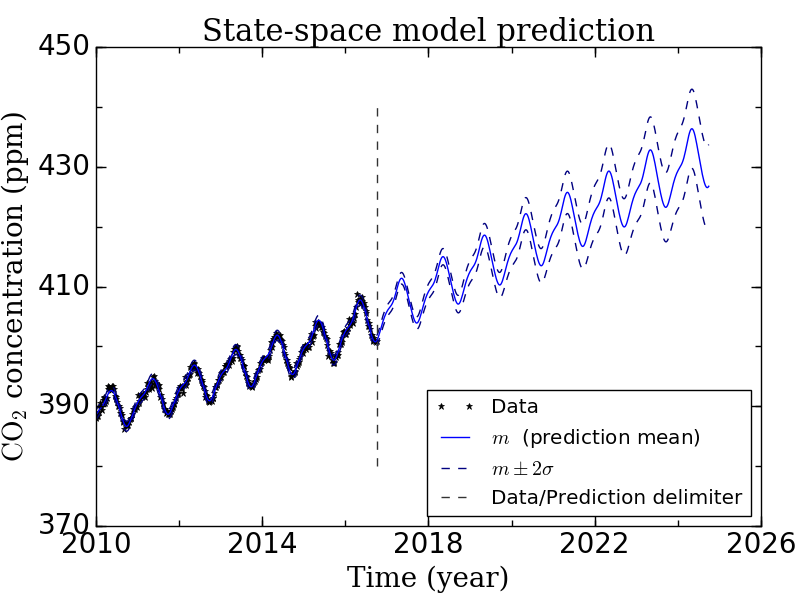}
        \caption{KF 8 years forecast}\label{fig:5c}
    \end{subfigure}
    \caption{Forecasting $CO_2$-concentration data}\label{fig:5}
\end{figure*}

The next test we have conducted is the scaling analysis of the same models with respect to precision matrix block sizes. From the same artificially generated data 1,000 data points have been taken and marginal likelihood and its derivatives have been calculated. Different kernels and kernel combinations have been tested so that block sizes of sparse covariance matrix change accordingly. The results are demonstrated in Fig.~\ref{fig:3}. In this figure, we see the superlinear scaling of SpInGP and KF inference ($O(b^3)$ in theory). The 1,000 data points is a relatively small number so the regular GP is much faster in this case.

\subsection{$CO_2$-Concentration Forecasting Experiment}

We have modeled and predicted the well-known real world dataset - atmospheric $CO_2$-concentration\footnote{Available at: ftp://ftp.cmdl.noaa.gov/ccg/ co2/trends/} measured at Mauna Loa Observatory, Hawaii. The detailed description of the data can be found e.g. in~\citep[p.~118]{rasmussen_2005}. It is a weekly sampled data starting on 5-th of May 1974 and ending on 2-nd of October 2016, in total 2195 measurements. The dataset is drawn on the Fig.~\ref{fig:5a}.

For the demonstration purpose rather intuitive covariance function is taken. It consist of 3 summands: quasi-periodic kernel models periodicity with possible long-term variations, Mat{\'e}rn$(\nu=\frac{3}{2})$ models short and middle term irregularities and Exponentiated-Quadratic (EQ) kernel models long-term trend:
\begin{equation}\label{eq:co2_1}
k(\cdot) = k_{\text{Quasi-Periodic}}^{\text{periodicity}}(\cdot) + k_{\text{Mat{\'e}rn$_{\frac{3}{2}}$}}^{\text{small variations}}(\cdot) + k_{\text{EQ}}^{\text{long trend}}(\cdot).
\end{equation}

The hyperparameters or the kernels have been optimized by running scaled conjugate gradient optimization
of marginal likelihood~\citep{rasmussen_2005}. The initial values of optimization have been chosen
to be preliminarily reasonable ones. %

From Fig.~\ref{fig:5b} and~\ref{fig:5c} it is noticeable that results are slightly different but sill very similar. The difference can be explained by slightly different optimal values of hyperparameters obtained during optimization and regularization applied to SpInGP.

\section{Conclusions}

In this paper we have proposed the sparse inverse formulation Gaussian process (SpInGP) algorithm for temporal Gaussian process regression. In contrast to standard Gaussian Processes which scales cubically with the number of time points, SpInGp scales linearly. Memory requirement is quadratic for standard GP and linear for SpInGP. And in contrast to inducing points sparse GPs SpInGP is exact for some kernels and arbitrary close to exact for others.

We have considered all computational subproblems which are encountered during the Gaussian process regression inference. There are four different subproblems which all can be solved by sequential or parallel algorithms for block-tridiagonal systems. Alternatively, the BTD formulation allows using general sparse solvers where the selective inversion operation is implemented. Hence, in contrast standard to Kalman filtering solution which scales linearly, our approach allows parallelization and sub-linear scaling.

\bibliography{../paper_bibliography}

\end{document}